\pdfoutput=1

\documentclass[11pt]{article}

\usepackage[]{acl}

\usepackage{times}
\usepackage{latexsym}

\usepackage[T1]{fontenc}

\usepackage[utf8]{inputenc}

\usepackage{microtype}

\usepackage{inconsolata}

\usepackage{amsmath}
\usepackage{amssymb}
\usepackage{mathtools}
\usepackage{amsthm}
\usepackage{enumitem}

\usepackage{multirow}
\usepackage{booktabs}
\usepackage{booktabs}
\usepackage{tabularx}
\usepackage{graphicx} 

\title{ShareLoRA: Parameter Efficient and Robust Large Language Model Fine-tuning via Shared Low-Rank Adaptation}

\author{
Yurun Song \\
UC Irvine \\
\texttt{yuruns@uci.edu} \\\And
 Junchen Zhao\\
UC Irvine \\
\texttt{junchez3@uci.edu} \\
\AND
Ian G. Harris \\
UC Irvine \\
\texttt{harris@ics.uci.edu} \\\And
Sangeetha Abdu Jyothi \\
UC Irvine, VMware Research \\
\texttt{sangeetha.aj@uci.edu} \\
}

\begin{document}
\maketitle

\begin{abstract}
In this paper, we introduce \textbf{Share}d \textbf{Lo}w \textbf{R}ank \textbf{A}daptation (ShareLoRA), a Large Language Model (LLM) fine-tuning technique that balances parameter efficiency, adaptability, and robustness without compromising performance. By strategically sharing the low-rank weight matrices across different layers, ShareLoRA achieves 44\% to 96\% reduction in trainable parameters compared to standard LoRA, alongside a substantial decrease in memory overhead. This efficiency gain scales with model size, making ShareLoRA particularly advantageous for resource-constrained environments. Importantly, ShareLoRA not only maintains model performance but also exhibits robustness in both classification and generation tasks across diverse models, including RoBERTa, GPT-2, and LLaMA series (1, 2, and 3).  It consistently outperforms LoRA in zero-shot, few-shot, and continual fine-tuning scenarios, achieving up to 1.2\% average accuracy improvement, and enhanced generalization across domains. In continual learning settings, ShareLoRA achieves 1.2\% higher accuracy on GSM8K, 0.6\% on HumanEval, and 0.5\% on both MMLU and MMLU-Pro.
Our results demonstrate that ShareLoRA supports high-quality fine-tuning while offering strong generalization and continual adaptation across various model scales and diverse tasks.\footnote{https://github.com/Rain9876/ShareLoRA}



\end{abstract}

\section{Introduction}
As Pretrained Language Models (PLMs) have gained prominence \cite{devlin2019bert, liu2019RoBERTa, Radford2019LanguageMA}, researchers are increasingly focused on optimizing the utilization of these models' pre-trained weights. Traditional fine-tuning, which involves adjusting all parameters of a PLM for a specific dataset or task, is often resource-intensive and time-consuming, especially given the massive scale of large language models (LLMs) \cite{brown2020language, kaplan2020scaling, hoffmann2022training, chowdhery2022palm, zhang2022opt, touvron2023LLaMA}.

Parameter-Efficient Fine-Tuning (PEFT) has proven to be an effective strategy for mitigating the challenges associated with extensive parameter adjustments. By modifying only a select subset of a model's parameters, PEFT enables cost-effective adaptation to domain-specific tasks while preserving performance levels comparable to those achieved with full fine-tuning \cite{pmlr-v97-houlsby19a, li-liang-2021-prefix, lin2020exploring, lei2023conditional, he2022unified, he2023mera, mahabadi2021parameterefficient}. Techniques like Low-Rank Adaptation (LoRA) \cite{hu2021LoRA} stand out within PEFT by demonstrating that models fine-tuned with a reduced parameter set can match the performance of those fine-tuned with full parameters, effectively bridging the gap in efficiency and efficacy.

Given the impressive performance of LoRA, subsequent studies have aimed to enhance its efficiency, mainly by reducing the number of trainable parameters to minimize the memory footprint during the fine-tuning process. However, significantly lowering the trainable parameters can lead to slow convergence, while insufficient reductions may encourage the model to easily overfit. Moreover, existing PEFT methods often struggle to maintain robustness across different domains after fine-tuning.


To address these challenges, we introduce ShareLoRA, an efficient and straightforward PEFT method that effectively balances trainable parameter selection while optimizing the model's adaptability, minimizing memory requirements, and ensuring robustness across domains. Our approach leverages the observation that low-rank weight matrices A and B do not need to be uniquely configured across layers to achieve optimal PEFT performance in PLMs. Instead, we propose sharing either matrix A or B across all layers while maintaining its counterpart as distinct in each layer. This strategy meets several key objectives: 
\begin{itemize}[leftmargin=*,nolistsep]
    \item \textbf{Parameter Efficiency:} Sharing a low-rank matrix across layers reduces trainable parameters by \textbf{44\%} to \textbf{96\%} compared to standard LoRA, for models such as LLaMA-7B. This memory reduction scales with model size which is critical for efficient fine-tuning LLMs on consumer GPUs and edge devices.
    \item \textbf{Model Adaptability:} Keeping the shared matrix trainable preserves the model's adaptability, allowing it to effectively learn and adapt to new tasks and domains. Also, the updated weights for each component that LoRA applies remain unique yet share a common base, promoting consistency across layers while allowing for task-specific adaptations.
    \item \textbf{Continual Adaption:} ShareLoRA exhibits robustness when continual fin-tuning to domains different from the one it was fine-tuned on. This generalization capability sets it apart from traditional LoRA and other PEFT methods, which often struggle to maintain performance when faced with out-of-domain tasks.
\end{itemize}

Our extensive experiments across multiple models, including RoBERTa, GPT-2, and LLaMA series, demonstrate that ShareLoRA not only preserves model performance but also shows remarkable robustness across a variety of tasks in both classification and generation. 


\begin{figure*}
  \centering 
\includegraphics[width=\textwidth]{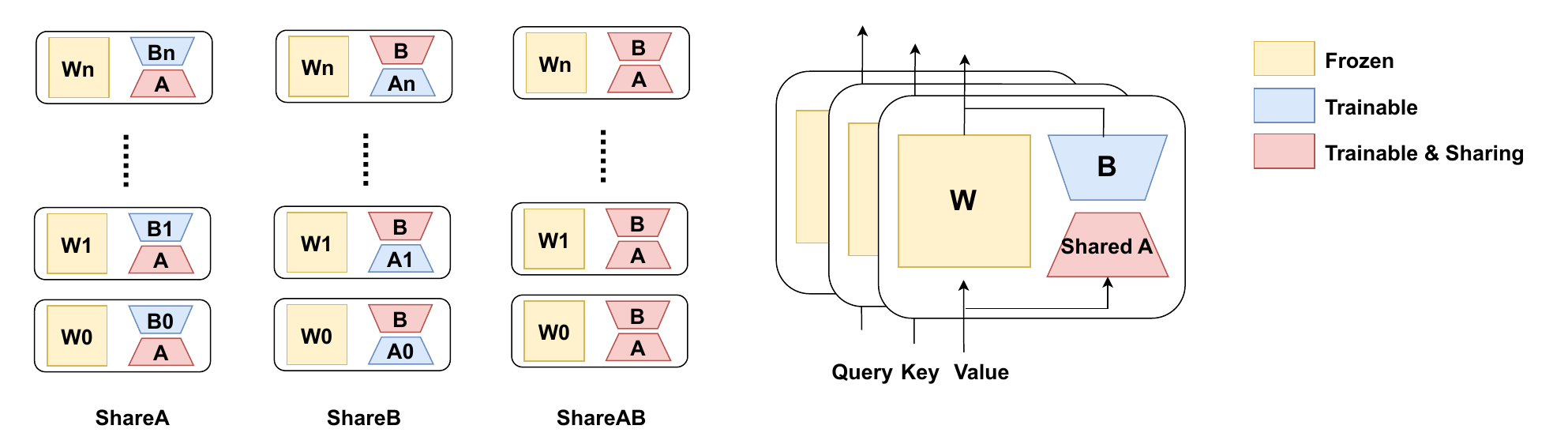}

  \caption{Overview of ShareLoRA: The implementation of ShareA, ShareB, and ShareAB across all layers (left), including ShareA applied across self-attention layers (right).}

  \label{fig:use_case}
\end{figure*}
\section{Related Works}
PLMs are trained on large datasets to develop broad linguistic representations \cite{devlin2019bert, liu2019RoBERTa, JMLR:v21:20-074}, but often fall short in specialized tasks due to a lack of domain knowledge. Traditional approaches involve fully fine-tuning PLMs to enhance domain-specific performance \cite{weiming2023, Jimmy2020, dabre-etal-2019-exploiting}. However, with the increasing size of PLMs \cite{workshop2023bloom, touvron2023LLaMA, touvron2023LLaMA2, zhang2022opt}, this method becomes too resource-heavy. As an alternative, Parameter Efficient Fine-tuning (PEFT) provides an efficient way to maintain performance with less computational expense.

PEFT methods have become crucial for adapting large-scale pre-trained models to specific tasks without extensively overhauling their parameters. This approach conserves computational resources and boosts efficiency. For example, Prefix tuning \cite{li-liang-2021-prefix} adds parameters to the hidden states across layers, subtly influencing the model's behavior without changing its underlying architecture. Prompt tuning \cite{lester-etal-2021-power} alters prompts and updates only the associated parameters, focusing on specific areas of model performance. BitFit \cite{zaken2022bitfit} updates only the biases within the model, resulting in minimal yet effective modifications.

One notable PEFT technique is Low-Rank Adaptation (LoRA) \cite{hu2021LoRA}, which achieves efficient fine-tuning by incorporating a low-rank matrix adaptation mechanism alongside the existing weights of linear layers. This approach reduces memory overhead while preserving the effectiveness of the fine-tuning process. 


Recent enhancements to LoRA have significantly broadened its capabilities. QLoRA \cite{dettmers2023qlora} optimizes LoRA for the fine-tuning of quantized models, thereby increasing efficiency. ReLoRA \cite{lialin2023relora} incorporates a warm-up strategy during pre-training to boost adaptability. LoraHub \cite{huang2024lorahub} streamlines the process by automating the creation of custom LoRA modules for specific tasks. Additionally, GLoRA \cite{chavan2023oneforall} introduces a prompt module that fine-tunes weights and biases, enhancing performance across a variety of applications.

Despite these advancements, LoRA still faces significant memory overhead due to high activation memory usage in LoRA layers during the fine-tuning phase. To address this issue, LoRA-FA \cite{zhang2023LoRAfa} strategically freezes the low-rank \( A \) matrix and updates only the \( B \) matrix. This approach significantly reduces the number of trainable parameters and activation memory, thus enhancing the efficiency of fine-tuning large language models without substantially impacting performance. 

However, LoRA-FA does not adequately decrease the total number of parameters that need to be stored, presenting a considerable challenge in contexts where computational resources and storage are constrained. Additionally, by freezing the \( A \) matrix, LoRA-FA limits the model's capacity to adapt and learn from new data during fine-tuning. This rigidity can hinder the model's performance, particularly in complex or domain-specific tasks. 

In contrast, our proposed approach ShareLoRA offers a more dynamic and flexible strategy by allowing either matrix \( A \) or \( B \), or both, to be shared across different layers. This method not only preserves the model's adaptability but also further reduces the memory requirements. 

\section{Method}
In this section, we provide a detailed description of our proposed PEFT approach ShareLoRA, as illustrated in Figure~\ref{fig:use_case}. ShareLoRA facilitates flexible configurations through two primary dimensions: \textbf{1.} the choice of sharing between the matrices A, B, or both A and B (ShareA, ShareB, and ShareAB), and \textbf{2.} the scope of sharing, which can be across different layers such as self-attention layers. This framework allows for a variety of combinations, enabling tailored adaptation of low-rank models to specific tasks.

\paragraph{ShareA Configuration}
In the ShareA configuration, the low-rank matrix \(A\) is uniformly shared across all layers, with each layer employing its own unique matrix \(B_i\). The formula for weight adaptation in each layer \(i\) can be expanded to detail the influence on model transformation:
\begin{equation}
    \Delta W_i = \alpha A B_i = \alpha \sum_{k=1}^{r} A_{:,k} B_{k,:,i}
\end{equation}
where \(A_{:,k}\) represents the \(k\)-th column of \(A\), and \(B_{k,:,i}\) is the \(k\)-th row of matrix \(B_i\). This equation shows that each layer's weight change, \(\Delta W_i\), is a linear combination of the columns of \(A\) weighted by the corresponding elements of \(B_i\). This shared projection-down matrix \(A\) reduces the dimensionality uniformly across all layers, thereby minimizing redundancy in learning and memory usage while enabling tailored output transformations through layer-specific matrices \(B_i\).

\begin{table*}[!ht]
\centering
\scalebox{0.7}{
\begin{tabularx}{1.28\textwidth}{l|c|ccccccccc}

\toprule
\textbf{Method} & \textbf{\# Params} & \textbf{MNLI} & \textbf{SST-2} & \textbf{MRPC} & \textbf{CoLA} & \textbf{QNLI} & \textbf{QQP} & \textbf{RTE} & \textbf{STS-B} & \textbf{Avg.}  \\ 
\midrule
R$_{\text{b}}$ (FT)* & 125.0M & \textbf{87.6} & 94.8 & 90.2 & 63.6 & 92.8 & \textbf{91.9} & 78.7 & 91.2 & 86.4 \\ 
R$_{\text{b}}$ (BitFit)* & 0.1M & 84.7 & 93.7 & \textbf{92.7} & 62.0 & 91.8 & 84.0 & 81.5 & 90.8 & 85.2 \\ 
R$_{\text{b}}$ (Adpt$^{\text{D}}$)* & 0.3M & $87.1_{\pm.0}$ & $94.2_{\pm .1}$ & $88.5_{\pm 1.1}$ & $60.8_{\pm .4}$ & $93.1_{\pm .1}$ & $90.2_{\pm 00}$ & $71.5_{\pm 2.7}$ & $89.7_{\pm .3}$ & 84.4 \\
R$_{\text{b}}$ (Adpt$^{\text{D}}$)* & 0.9M & $87.3_{\pm .1}$ & $94.7_{\pm .3}$ & $88.4_{\pm .1}$ & $62.6_{\pm.9}$ & $93.0_{\pm.2}$ & $90.6_{\pm .0}$ & $75.9_{\pm 2.2}$ & $90.3_{\pm .1}$ & 85.4 \\
R$_{\text{b}}$ (Prefix)* & 0.36M & $85.21$ & $93.81$ & $87.25$ & $59.31$ & $90.77$ & $87.75$ & $54.51$ & $88.48$ & 80.9 \\
R$_{\text{b}}$ (IA$^3$)* & 0.06M & $83.95$ & $93.92$ & $87.00$ & $59.58$ & $90.88$ & $87.99$ & $71.12$ & $90.30$ & 83.1 \\
R$_{\text{b}}$ (LoRA)* & 0.3M & $87.5_{\pm .3}$ & $\mathbf{95.1_{\pm .2}}$ & $89.7_{\pm .7}$ & $63.4_{\pm 1.2}$ & $\mathbf{93.3_{\pm .3}}$ & $90.8_{\pm .1}$ & $86.6_{\pm .7}$ & $\mathbf{91.5_{\pm .2}}$ & \textbf{87.2} \\
R$_{\text{b}}$ (L-FA)* & 0.15M & $86.8$ & $94.8$ & $90$ & $63.6$ & $92.5$ & $90.1$ & $67.9$ & $89.6$ & 84.4 \\
R$_{\text{b}}$ (VERA)* & 0.04M & --- & $94.6_{\pm .1}$ & $89.5_{\pm .5}$ & $65.6_{\pm .8}$ & $91.8_{\pm .2}$ & --- & $78.7_{\pm .7}$ & $90.7_{\pm .2}$ & 85.2 \\
R$_{\text{b}}$ (Tied-LoRA)* & 0.04M & --- & $94.4_{\pm .5}$ & $88.5_{\pm 1.0}$ & $61.9_{\pm 1.6}$ & $92.2_{\pm .2}$ & --- & $76.2_{\pm 1.0}$ & $89.8_{\pm .3}$ & 83.8 \\
R$_{\text{b}}$ (VB-LoRA)* & 0.03M & --- & $94.4_{\pm .2}$ & $89.5_{\pm .5}$ & $63.3_{\pm .7}$ & $92.2_{\pm .2}$ & --- & $82.3_{\pm 1.3}$ & $90.8_{\pm .1}$ & 85.4 \\

R$_{\text{b}}$ (ShareA) & 0.16M & $87.3_{\pm .2}$ & $95.0_{\pm .3}$ & $89.9_{\pm .8}$ & $\mathbf{63.8_{\pm 1.1}}$ & $92.8_{\pm .18}$ & $90.3_{\pm .05}$ & $\mathbf{87.1_{\pm .5}}$ & $91.4_{\pm .1}$ & \textbf{87.2} \\
\midrule \midrule
R$_{\text{l}}$ (FT)* & 335.0M & $90.2$ & $\mathbf{96.4}$ & $90.9$ & $68.0$ & $94.7$ & $\mathbf{92.2}$ & $86.6$ & $92.4$ & 88.9 \\ 
R$_{\text{l}}$ (LoRA)* & 0.8M & $90.6_{\pm .2}$ & $96.2_{\pm .5}$ & $90.9_{\pm 1.2}$ & $\mathbf{68.2_{\pm 1.9}}$ & $94.9_{\pm .3}$ & $91.6_{\pm .1}$ & $87.4_{\pm 1.1}$ & $\mathbf{92.6_{\pm .2}}$ & 89.0 \\
R$_{\text{l}}$ (L-FA)* & 0.4M & $90.1$ & $96$ & $90$ & $68$ & $94.4$ & $91.1$ & $86.1$ & $92$ & 88.5 \\ 
R$_{\text{l}}$ (VeRA)*      & 0.06M & --- & $96.1_{\pm0.1}$ & $90.9_{\pm0.7}$ & $68.0_{\pm0.8}$ & $94.4_{\pm0.2}$ & --- & $85.9_{\pm0.7}$ & $91.7_{\pm0.8}$ & 87.8  \\
R$_{\text{l}}$ (Tied-LoRA)* & 0.07M & --- & $94.8_{\pm0.6}$ & $89.7_{\pm1.0}$ & $64.7_{\pm1.2}$ & $94.1_{\pm0.1}$ & --- & $81.2_{\pm0.1}$ & $90.8_{\pm0.3}$ & 85.9 \\
R$_{\text{l}}$ (VB-LoRA)*   & 0.03M & --- & $96.1_{\pm0.2}$ & $\mathbf{91.4_{\pm0.6}}$ & $68.3_{\pm0.7}$ & $94.7_{\pm0.5}$ & --- & $86.6_{\pm1.3}$ & $91.8_{\pm0.1}$ & 88.2 \\
R$_{\text{l}}$ (ShareA) & 0.4M & $\mathbf{90.7_{\pm .1}}$ & $96.1_{\pm .1}$ & $91.1_{\pm .8}$ & $67.7_{\pm 1.5}$ & $\mathbf{95.1_{\pm .1}}$ & $91.3_{\pm .1}$ & $\mathbf{90.3_{\pm .3}}$ & $92.5_{\pm .1}$ & \textbf{89.3}\\
\midrule
R$_{\text{l}}$ (Prefix)* & 0.9M & $89.30$ & $95.76$ & $88.24$ & $59.01$ & $93.32$ & $88.88$ & $74.01$ & $90.92$ & 84.9 \\
R$_{\text{l}}$ (IA$^3$)* & 0.18M & $88.63$ & $94.61$ & $86.52$ & $61.15$ & $94.25$ & $89.45$ & $81.23$ & $92.22$ & 86.0 \\
R$_{\text{l}}$ (LoRA)$\dag$ & 0.8M & $90.6_{\pm .2}$ & $\mathbf{96.2_{\pm .5}}$ & $90.2_{\pm 1.0}$ & $\mathbf{68.2_{\pm 1.9}}$ & $94.8_{\pm .3}$ & $\mathbf{91.6_{\pm .2}}$ & $85.2_{\pm 1.1}$ & $\mathbf{92.3_{\pm .5}}$ & \textbf{88.6} \\

R$_{\text{l}}$ (ShareAB)$\dag$ & 0.03M & $90.2_{\pm .1}$ & $95.9_{\pm .3}$ & $89.7_{\pm 1.0}$ & $62.3_{\pm .9}$ & $94.6_{\pm .1}$ & $89.7_{\pm .1}$ & $83.0_{\pm 0.8}$ & $90.3_{\pm .2}$ & 87.0 \\
R$_{\text{l}}$ (ShareB)$\dag$ & 0.4M & $90.4_{\pm .1}$ & $96.0_{\pm .3}$ & $\mathbf{90.4_{\pm .4}}$ & $65.8_{\pm .8}$ & $94.6_{\pm .1}$ & $91.0_{\pm .1}$ & $84.1_{\pm 1.2}$ & $91.4_{\pm .2}$ & 88.0 \\
R$_{\text{l}}$ (ShareA)$\dag$ & 0.4M & $\mathbf{90.7_{\pm .1}}$ & $96.1_{\pm .1}$ & $90.0_{\pm .5}$ & $67.7_{\pm 1.5}$ & $\mathbf{95.0_{\pm .1}}$ & $91.3_{\pm .1}$ & $\mathbf{85.9_{\pm .8}}$ & $91.8_{\pm .2}$ & \textbf{88.6} \\

\bottomrule 
\end{tabularx}
}
\caption{RoBERTa$_{base}$ and RoBERTa$_{large}$ with different adaptation methods on the GLUE benchmark. $*$ indicates numbers published in prior works. $\dag$ indicates runs configured in a setup similar to \cite{pmlr-v97-houlsby19a} and \cite{hu2021LoRA} for a fair comparison. \label{GLUE}}
\end{table*}

\paragraph{ShareB Configuration}
In the ShareB configuration, matrix \(B\) is uniformly shared across all layers, while each layer employs its own unique matrix \(A_i\). The weight adjustment for each layer is expressed as:
\begin{equation}
    \Delta W_i = \alpha A_i B = \alpha \sum_{k=1}^{r} A_{i,:,k} B_{k,:}
\end{equation}
where \(A_{i,:,k}\) denotes the \(k\)-th column of matrix \(A_i\) for layer \(i\), and \(B_{k,:}\) represents the \(k\)-th row of the shared matrix \(B\). Here, the uniform projection-up matrix \(B\) ensures consistent expansion of the transformed data back to the output dimension across all layers, while the distinct \(A_i\) matrices allow for adaptation to the specific input characteristics of each layer.

\paragraph{ShareAB Configuration}
When both matrices \(A\) and \(B\) are shared across all layers, the change in weights is simplified, leading to substantial parameter reduction:
\begin{equation}
    \Delta W = \alpha AB = \alpha \sum_{k=1}^{r} A_{:,k} B_{k,:}
\end{equation}
where both \(A_{:,k}\) and \(B_{k,:}\) are shared across all layers. This configuration significantly reduces the model complexity by eliminating the need for distinct matrices in each layer, thus reducing memory requirements and computational overhead. The entire model operates under a uniform transformation schema, which simplifies training and storage but requires careful calibration of the initial values and ongoing adjustments during fine-tuning to preserve model effectiveness across diverse tasks.

\paragraph{Sharing Across Self-Attention Layers}
In the ShareA configuration of ShareLoRA applied to PLMs across all self-attention layers, the matrices \(A_Q\), \(A_K\), and \(A_V\) are shared. These matrices are responsible for reducing the dimensionality of the inputs for Queries (Q), Keys (K), and Values (V) respectively, we term it as \textbf{ShareA$_{qkv}$} in the following paragraphs. The process for each component in the \(i\)-th self-attention layer is formalized as follows:
\begin{align}
    Q_i &= X_i A_Q B_{Q_i} \\
    K_i &= X_i A_K B_{K_i} \\
    V_i &= X_i A_V B_{V_i} \\
    \text{Attention}(Q_i, K_i, V_i) &= \text{softmax}\left(\frac{Q_i K_i^T}{\sqrt{d_{K_i}}}\right) V_i,
\end{align}
where \(X_i\) denotes the input to the \(i\)-th self-attention layer. Each matrix \(A_Q\), \(A_K\), and \(A_V\) facilitates a consistent reduction in input dimensions across all layers, which simplifies the model architecture by maintaining a uniform approach to processing the foundational aspects of self-attention. The unique matrices \(B_{Q_i}\), \(B_{K_i}\), and \(B_{V_i}\) for each component allow for tailored transformations that meet the specific needs of each self-attention layer.

\begin{table*}[!ht]
\centering
\scalebox{0.8}{
\begin{tabularx}{0.95\textwidth}{l|c|ccccc}
\toprule
\textbf{Method} & \textbf{\# Params} & \textbf{BLUE} & \textbf{NIST} & \textbf{MET} & \textbf{ROUGE-L} & \textbf{CIDEr} \\ 
\midrule
GPT-2 M (FT)* & 354.92M & 68.2 & 8.62 & 46.2 & 71.0 & 2.47 \\ 
GPT-2 M (AdapterL)* & 0.37M & 66.3 & 8.41 & 45.0 & 69.8 & 2.40  \\ 
GPT-2 M (AdapterL)* & 11.09M & 68.9 & 8.71 & 46.1 & 71.3 & 2.47 \\ 
GPT-2 M (PreLayer)* & 0.35M & 69.7 & \textbf{8.81} & 46.1 & 71.4 & 2.49  \\ 
GPT-2 M (VeRA)* & 0.10M & $\mathbf{70.1}$ & $\mathbf{8.81}$ & $46.6$ & $71.5$  &$2.50$  \\ 
GPT-2 M (LoRA) & 0.35M & $69.5_{\pm .7}$ & $8.74_{\pm .08}$ & $46.56_{\pm .2}$  & $71.51_{\pm .3}$ &$2.50_{\pm .01}$ \\ 
GPT-2 M (ShareB) & 0.20M & $67.1_{\pm .7}$ & $8.55_{\pm .09}$ & $45.12_{\pm .4}$ & $69.45_{\pm .6}$  &$2.37_{\pm .01}$  \\ 
GPT-2 M (ShareA) & 0.20M & $69.7_{\pm .4}$ & $8.75_{\pm .05}$ & $\mathbf{46.60_{\pm .1}}$ & $\mathbf{71.63_{\pm .1}}$  &$\mathbf{2.51_{\pm .01}}$  \\ 
\midrule
GPT-2 L (FT)* & 774.03M & 68.5 & 8.78 & 46.0 & 69.9 & 2.45 \\ 
GPT-2 L (AdapterL)* & 0.88M & 69.1 & 8.68 & 46.3 & 71.4 & 2.49  \\ 
GPT-2 L (AdapterL)* & 23.00M & 68.9 & 8.70 & 46.1 & 71.3 & 2.45 \\ 
GPT-2 L (PreLayer)* & 0.77M & \textbf{70.3} & \textbf{8.85} & 46.3 & 71.7 & 2.47  \\ 
GPT-2 L (VERA)* & 0.17M & $\textbf{70.3}$ & $\textbf{8.85}$ & $\textbf{46.9}$ & $71.6$  &$2.52$  \\
GPT-2 L (LoRA) & 0.77M & $69.8_{\pm .4}$ & $8.80_{\pm .04}$ & $46.69_{\pm .1}$  & $71.71_{\pm .3}$ &$\mathbf{2.52_{\pm .01}}$ \\ 
GPT-2 L (ShareB) & 0.39M & $69.7_{\pm .2}$ & $8.80_{\pm .01}$ & $46.17_{\pm .3}$ & $70.94_{\pm .5}$  &$2.49_{\pm .02}$  \\ 
GPT-2 L (ShareA) & 0.39M & $70.0_{\pm .1}$ & $8.83_{\pm .03}$ & $46.60_{\pm .1}$ & $\mathbf{71.74_{\pm .1}}$  &$\mathbf{2.52_{\pm .02}}$  \\ 
\bottomrule
\end{tabularx}
}
\caption{GPT-2 medium (M) and large (L) with different adaptation methods on the E2E NLG
Challenge. For all metrics, higher is better. LoRA ShareA outperforms several baselines with comparable
or fewer trainable parameters. * indicates numbers published in prior works.} \label{e2e}
\end{table*}

\section{Experiments}
 In our study, we conduct a comprehensive evaluation of the downstream performance of ShareLoRA across several series models, including RoBERTa \cite{liu2019RoBERTa} and GPT-2 \cite{Radford2019LanguageMA}. We benchmark these results against other established approaches such as LoRA \cite{hu2021LoRA}, LoRA-FA \cite{zhang2023LoRAfa}. Additionally, we extend the application of ShareLoRA to large-scale model in LLaMA series (\citealp{touvron2023LLaMA}, \citealp{touvron2023LLaMA2}, \citealp{dubey2024llama}) architectures, particularly in few-shot, zero-shot scenarios. Furthermore, our experiments cover a range of model sizes, from 7 billion to 13 billion parameters, and included both quantized and unquantized model variants. All tests were performed on the Nvidia A6000 and RTX 3090 GPUs. For experiment hyper-parameter settings, see Appendix Table~\ref{tab:training_details1}-Table~\ref{tab:parameters2}.


\begin{table*}[!ht]
\centering
\scalebox{0.7}{
\begin{tabularx}{1.1\textwidth}{l|c|c||l|c|c}
\toprule
\textbf{Method} & \textbf{\# Params} & \textbf{MMLU} & \textbf{Method} & \textbf{\# Params} & \textbf{MMLU} \\ 
\midrule
LLaMA 7B *                     & 6738.4M  & 35.1   & LLaMA 13B  *                     & 13015M   & 46.9  \\ 
LLaMA 7B (LoRA)*               & 159.9M   & 40.67  & LLaMA 13B (LoRA)*               & 250.3M   & 47.49  \\
LLaMA 7B (LoRA)                & 159.9M   & $\mathbf{41.65_{\pm1.0}}$  & LLaMA 13B (LoRA)                & 250.3M   & $47.60_{\pm1.4}$  \\
LLaMA 7B (ShareA$_{qkv}$)      & 135.5M   & $41.01_{\pm0.8}$  & LLaMA 13B (ShareA$_{qkv}$)      & 212.0M   & $\mathbf{48.76_{\pm0.7}}$  \\ 
LLaMA 7B (ShareA)              & 89.3M    & $40.93_{\pm0.5}$   & LLaMA 13B (ShareA)              & 139.1M   & $48.15_{\pm0.5}$  \\ 
\hline\midrule
LLaMA2 7B *                    & 6898.3M  & 45.7            & LLaMA2 13B *                     & 13266M  & 53.8 \\ 
LLaMA2 7B (LoRA)               & 159.9M   & $47.47_{\pm1.1}$ & LLaMA2 13B (LoRA)               & 250.3M  & $55.31_{\pm0.2}$ \\ 
LLaMA2 7B (ShareA$_{qkv}$)     & 135.5M   & $47.88_{\pm0.1}$ & LLaMA2 13B (ShareA$_{qkv}$)     & 212.0M  & $\mathbf{55.66_{\pm0.1}}$  \\ 
LLaMA2 7B (ShareA)             & 89.3M    & $\mathbf{48.19_{\pm0.4}}$   & LLaMA2 13B (ShareA)  & 139.1M   & $55.53_{\pm0.3}$  \\ 
\bottomrule
\end{tabularx}
}
\caption{LLaMA and LLaMA2, ranging from 7B to 13B, are fine-tuned using different sharing approaches on the Alpaca datasets and evaluated on the MMLU 5 shot benchmark. The configuration runs is based on the setup described in \cite{dettmers2023qlora}.* indicates numbers published in prior works, reported by \cite{xu2023parameterefficient}.}\label{MMLU}
\end{table*}

\subsection{Datasets}
The experiment datasets are primarily divided into three categories: Natural Language Understanding (NLU), Natural Language Generation (NLG) and few-shot tasks, using the same configuration and datasets as LoRA \cite{hu2021LoRA} and \cite{dettmers2023qlora}. 

For NLU, we employ the GLUE benchmark \cite{wang2019glue}, which includes MNLI, SST-2, MRPC, CoLA, QNLI, QQP, RTE, and STS-B tasks. Notably, for MRPC, RTE, and STS-B tasks, we initialize the LoRA modules with the trained MNLI checkpoint as \cite{hu2021LoRA} demonstrated.
For NLG, we replicate experiments similar to those of LoRA using the E2E challenge dataset \cite{novikova2017e2e}, following the same experimental setup.

Additionally, we expand our experiments to few-shot and zero-shot tasks on larger models, demonstrating our approach's adaptability. Following the configuration outlined in \cite{dettmers2023qlora}, we employ Alpaca \cite{alpaca}, CodeAlpaca\cite{codealpaca} and MATH \cite{hendrycksmath2021} for LoRA and ShareLoRA, using the MMLU benchmark ~\cite{hendrycks2021measuring} for evaluation. Some other benchmarks like ARC~\cite{arc}, Hellaswrag~\cite{hellaswag}, MMLU-Pro \cite{wang2024mmlupro}, HumanEval\cite{chen2021codex} and GSM8K~\cite{gsm8k} are used for comparison of model adaptability.
All experimental setups are consistent with those described studies and demonstration of their repositories, based on the best of our knowledge.

\subsection{Baselines}
\textbf{Full Fine-Tuning (FT)}
is a commonly used approach for model adaptation involving with updating all model's parameters.\\
\textbf{LoRA}
\cite{hu2021LoRA} is a technique that introduces a pair of rank decomposition trainable matrices alongside existing weight matrices in neural networks.\\
\textbf{Bitfit}~\cite{zaken2022bitfit} is a technique  for updating only a select small subset of biases parameters, to improve performance on new tasks while freezing all other pre-trained weights.\\
\textbf{PreLayer/Prefix} \cite{li2021prefixtuning} is a parameter-efficient technique for customizing large language models by learning specific activations after each Transformer layer for designated prefix tokens, while the main model parameters remain unchanged.\\ 
\textbf{Adapter} \cite{pmlr-v97-houlsby19a} involves inserting adapter layers between neural modules such as the self-attention and MLP modules, enhancing model flexibility without extensive modifications.
AdapterL \cite{lin2020exploring} introduces adapters after the MLP module followed by a LayerNorm, while AdapterD \cite{rücklé2021adapterdrop} increases efficiency by omitting some adapter layers. \\
\textbf{IA$^3$} \cite{liu2022few} is a PEFT approach that enhances model performance by scaling activations with learned vectors. \\
\textbf{LoRA-FA}
\cite{zhang2023LoRAfa} is a memory-efficient approach to fine-tuning large language models by reducing the activation memory required. \\
\textbf{VERA}
\cite{kopiczko2023vera} reduces trainable parameters by using frozen random matrices and learned scaling vectors, matching LoRA’s performance more efficiently.\\
\textbf{Tied-LoRA}
\cite{renduchintala2023tied} improves parameter efficiency by tying weights and training fewer low-rank matrices, matching LoRA performance with significantly fewer parameters.\\
\textbf{VB-LoRA}
\cite{li2024vb} achieves extreme parameter efficiency by generating low-rank adaptation weights from a shared vector bank using a differentiable top-k selection.\\

\begin{table}[!ht]
\centering
\scalebox{0.65}{
\begin{tabularx}{1.5\linewidth}{l|cccc}
\toprule
\textbf{Method}     & \textbf{MMLU} & \textbf{ARC (c)} & \textbf{Hellaswarg} & \textbf{GSM8K} \\
\midrule
LLaMA 7B (LoRA)       & 41.28 & 48.49    & 76.74     & 2.43 \\
LLaMA 7B (ShareA)     & 40.67 & 48.82    & 76.67     & 3.16 \\
LLaMA 13B (LoRA)      & 45.02 & \textbf{51.34}    & 79.46     & 5.79 \\
LLaMA 13B (ShareA)    & \textbf{46.04} & 51.19    & \textbf{79.53}     & \textbf{6.17} \\ 
\midrule \midrule
LLaMA2 7B (LoRA)      & 45.68 & 49.60   &  77.14    & 3.21  \\ 
LLaMA2 7B (ShareA)    & 47.09 & 50.14   &  76.77    & 6.06  \\
LLaMA2 13B (LoRA)     & 53.21 & 51.28   &  76.59    & 12.33 \\
LLaMA2 13B (ShareA)   & \textbf{53.70} & \textbf{52.48}   &  \textbf{79.43}    & \textbf{14.99} \\ 
\bottomrule
\end{tabularx}
}
\caption{Performance of LLaMA models trained on the Alpaca General dataset and tested in a zero-shot of MMLU, ARC Challenge, and Hellaswarg, and in a five-shot of GSM8K, using the lm-eval-harness leaderboard \cite{eval-harness}. This table demonstrates the model's cross-domain adaptability in common sense, reasoning, and mathematics after finetuning on the general dataset.}
\label{lm_eval1}
\end{table}

\section{Results}
\paragraph{Parameter Efficiency and Performance} ShareLoRA demonstrates significant parameter efficiency while maintaining or improving performance across various model sizes and tasks. For large-scale LLaMA models, as shown in Table~\ref{MMLU}, ShareA reduces trainable parameters by \textbf{44\%} compared to LoRA. Despite this substantial reduction, ShareA achieves comparable or improved MMLU scores, with LLaMA 13B showing an increase from 47.60 to 48.15.



On the E2E NLG Challenge in Table~\ref{e2e}, ShareA demonstrates markedly greater efficiency on GPT-2 models: it reduces LoRA’s parameter count by 43\% on the Medium model, yet still achieves performance gains. Specifically, GPT-2 Medium’s BLEU improves from 69.5 to 69.7 and its ROUGE-L from 71.51 to 71.63, while GPT-2 Large’s BLEU increases from 69.8 to 70.0.

Notably, while ShareA consistently outperforms LoRA, our experiments show that ShareB and ShareAB generally underperform compared to ShareA. For instance, in the GPT-2 Large model, Table~\ref{e2e} shows ShareB achieves a BLEU score of 69.7 and a ROUGE-L score of 70.94, which are lower than both LoRA and ShareA. 

Comparing ShareLoRA to other state-of-the-art PEFT methods, we observe competitive or superior performance. For instance, on the GLUE benchmark using RoBERTa-large, Table~\ref{GLUE} shows ShareA achieves an average score of 88.6 on GLUE, compared to 84.9 for Prefix-tuning while using significantly fewer parameters. Even in its most aggressive configuration, ShareAB, with only 0.03M trainable parameters which reduces \textbf{96\%} trainable parameters compared to LoRA, outperforms IA$^3$ which uses 0.18M parameters, achieving an average score of 87.0 on GLUE compared to IA$^3$'s 86.0. Furthermore, under a similar trainable parameter size, ShareA demonstrates better performance than LoRA-FA. For example, ShareA achieves an average GLUE score of 89.3 with 0.4M parameters on RoBERTa-large, surpassing LoRA-FA's score of 88.5 with the same parameter count.

\paragraph{Model Adaptability} ShareLoRA demonstrates superior adaptability across a diverse range of tasks and model sizes. In experiments with RoBERTa-base model on the GLUE benchmark shown in Table~\ref{GLUE}, ShareA exhibits particular strength on smaller datasets that are typically prone to overfitting. Specifically, on tasks such as MRPC, CoLA, and RTE, ShareA achieves performance gains of 0.2\% to 0.5\%. These improvements are especially noteworthy given that these datasets have generally reached full convergence under standard training configurations \cite{hu2021LoRA}, suggesting ShareLoRA's ability to extract additional performance even in challenging scenarios.

\begin{table}[!ht]
\scalebox{0.75}{
\begin{tabularx}{1.3\linewidth}{l|cccc}
\toprule
\textbf{Method}        & \textbf{MATH} & \textbf{MMLU} & \textbf{MMLU Pro}\\
\midrule
LLaMA3 8B (LoRA)       & 12.46    & 65.93     & 31.80   \\
LLaMA3 8B (ShareA)     & \textbf{13.24}    & \textbf{66.35}     & \textbf{32.55}   \\
\midrule \midrule
LLaMA3.1 8B (LoRA)     & \textbf{15.36}    & 65.59     & 32.86   \\
LLaMA3.1 8B (ShareA)   & 15.10    & \textbf{65.78}     & \textbf{33.69}   \\ 
\bottomrule
\end{tabularx}}
\caption{Performance of LLaMA3 trained on the MATH dataset \citep{hendrycksmath2021} and evaluated in a zero-shot of MATH and in five-shot of MMLU and MMLU-Pro. It highlights the model's ability to maintain cross-domain adaptability in common sense and reasoning domains after finetuning on mathematics.}
\label{lm_eval2}
\end{table}

\begin{table*}[!ht]
\centering
\scalebox{0.88}{
\begin{tabular}{l|cc|c|c|cc}
\hline\toprule
\textbf{LLaMA3 8B} & \multicolumn{2}{c|}{\textit{Phase 1} \textbf{ALPACA}}      & \textit{Phase 2} \textbf{GSM8K} & \textit{Phase 3} \textbf{CodeALPACA} & \multicolumn{2}{c}{\textit{Phase 4} \textbf{ALPACA}}                           \\ \hline
Tasks & \multicolumn{1}{c}{\textbf{MMLU}}     & \textbf{MMLU Pro} & \textbf{GSM8K}          & \textbf{HumanEval}           & \multicolumn{1}{c}{\textbf{MMLU}}     & \multicolumn{1}{c}{\textbf{MMLU Pro}} \\ \hline
LoRA$_{qv}$       & \multicolumn{1}{c}{65.44} & 33.15   & 55.64        & 38.41             & \multicolumn{1}{c}{\textbf{65.32}} & 33.14                       \\ 
ShareLoRA$_{qv}$  & \multicolumn{1}{c}{\textbf{65.94}} & \textbf{33.85}   & \textbf{56.78}        & \textbf{39.02}             & \multicolumn{1}{c}{\textbf{65.30}} & \textbf{33.36}                       \\ \bottomrule
\end{tabular}}
\caption{Continual Adaption across multiple tasks, starting with Alpaca, followed by GSM8K, then CodeAlpaca, and finally revisiting Alpaca. At each stage, we evaluate the effectiveness of continual adaptation by leveraging the best checkpoints from the preceding task, comparing both LoRA and ShareLoRA for LLaMA3 8B.}\label{continual}
\end{table*}


\begin{figure}[ht!]
  \centering 
  \includegraphics[width=0.95\linewidth]{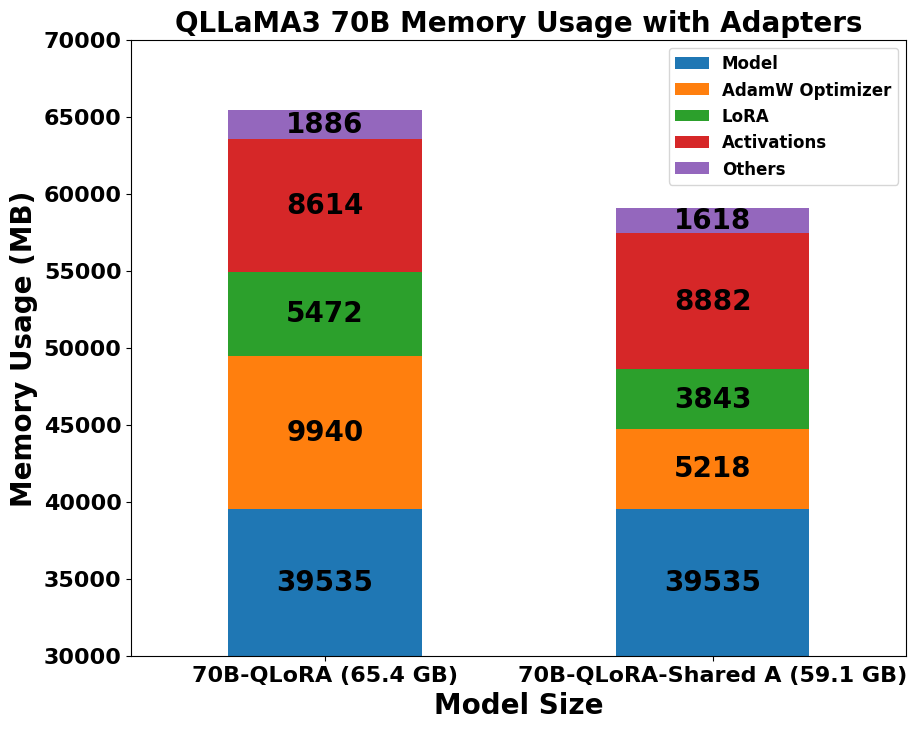} 
  \caption{Memory Consumption of LLaMA3 70B with QLoRA and QLoRA-shareA (QShareA).}
  \label{fig:peak}
\end{figure}

ShareA further showcases enhanced transfer learning capabilities. When fine-tuning on adaptive tasks like MRPC, RTE, and STS-B using the best MNLI checkpoint, ShareA consistently performs on par with or outperforms LoRA. Notably, ShareA outperforms other PEFT methods in this transfer learning scenario as well. For instance, on the RTE task, ShareA, with 0.16M parameters for RoBERTa-base, achieves a score of 87.1, significantly surpassing Prefix-tuning's 54.51 as shown in Table~\ref{GLUE}. ShareA also demonstrates superior performance when compared to methods with similar trainable parameter sizes, such as BitFit with 0.1M parameters and LoRA-FA with 0.15M parameters. This highlights ShareA's efficiency in parameter utilization and its ability to extract better performance from a given parameter budget, particularly in transfer learning scenarios.

\paragraph{Robustness Across Domains}

ShareLoRA shows strong robustness and adaptability across both diverse task domains and varying model sizes. As presented in Tables~\ref{MMLU} and~\ref{lm_eval1}, ShareLoRA consistently surpasses LoRA in zero-shot and few-shot learning scenarios across multiple evaluation benchmarks.

On the LLaMA2-7B model, ShareLoRA improves MMLU accuracy by 0.7\%, while on the LLaMA2-13B model, it achieves a 0.5\% gain. Beyond MMLU, ShareLoRA delivers average performance gains of 1.8\% and 1.3\% on LLaMA2-7B and LLaMA2-14B models, respectively, with accuracy improvements ranging from 0.5\% to 2.5\% across various tasks. These results collectively underscore ShareLoRA’s effectiveness in enhancing model generalization and transferability across both small and large-scale language models.

\paragraph{Continual Adaptation}
To assess the robustness and knowledge retention during continual fine-tuning, we deploy the LLaMA3 and LLaMA3.1 models on the MATH dataset. We then evaluate their performance in mathematics and across other domains, such as MMLU and MMLU-Pro, to compare how well these models preserve knowledge, as shown in Table~\ref{lm_eval2}. 
Our findings indicate that both ShareLoRA and LoRA deliver matched performances for directly fine-tuned domains. However, when adapting these fine-tuned models to other evaluation benchmarks, ShareLoRA demonstrates greater robustness, outperforming LoRA. Specifically, on MMLU-Pro, ShareLoRA outperforms LoRA by 0.86\% on LLaMA3.1 and 0.75\% on LLaMA3.

We also investigate continual fine-tuning across multiple tasks—starting from Alpaca, followed by GSM8K, then CodeAlpaca, and finally returning to Alpaca in Table~\ref{continual}. ShareLoRA consistently outperforms LoRA in this setting, with observed gains of 0.5\% on MMLU and MMLU-Pro, 1.2\% on GSM8K, and 0.6\% on HumanEval, highlighting its robustness in multi-task continual learning.



\section{Analysis and Discussion}

\paragraph{Relative Importance of LoRA Components}
Our experimental findings demonstrate that both LoRA and ShareA consistently outperform ShareB in a variety of classification and generative tasks, across most metrics. Within the LoRA framework, the up-projection matrix B plays a pivotal role by significantly enhancing the dimensionality of the low-rank representation. Consequently, it is both practical and justifiable to share the less critical module, LoRA A, while retaining the integrity of B. However, sharing both matrices A and B simultaneously tends to compromise too much critical information. Particularly in generative tasks, opting to share component A instead of B within the ShareLoRA framework is strategically beneficial, as seen in Table \ref{e2e}. This is because expanding the intermediate dimension proves more crucial and challenging than compressing high-dimensional features in complex generative scenarios.

\paragraph{Sharing Attention QKV vs. Sharing All} \label{QKV} 
The distinction between sharing the self-attention mechanism and all linear modules exists on MLP components like gates and up/down projections. This leads to a discrepancy in trainable parameters between LoRA's A and B. The strategic choice involves deciding whether to uniformly share weights across all layers (ShareA) or selectively share them, such as only for the down projection (ShareAB) while maintaining unique weights for other components like the up projection and gates. Preliminary results in Appendix Figure~\ref{fig:plot} suggest that selective sharing, particularly of the QKV matrices in Share$_{qkv}$, provides an effective balance by aligning closely with both ShareA and LoRA , potentially mitigating overfitting risks.

\paragraph{Memory Footprint}
In the context of smaller models like RoBERTa and GPT-2, ShareA yields minimal parameter savings, which is negligible given modern GPU capacities. However, with larger models like LLaMA, ShareA demonstrates more substantial reductions. Specifically, the LLaMA 7B and 13B models cut down approximately 60 million and 110 million trainable parameters, when compared to the LoRA. This leads to substantial efficiency gains, reducing both computational footprint and disk storage needs.

As depicted in Figure~\ref{fig:peak} and Figure~\ref{fig:memory_size},  in the Llama3 70B model, the ShareA adaptation achieves a 6.3GB reduction in memory footprint under the quantization configuration. Meanwhile, in the Llama2 13B model with the LoRA configuration, ShareA manages to reduce the memory footprint by 3.8GB and enhances training speed by approximately 3\%.  The confidence intervals in Table~\ref{MMLU} illustrate that ShareA not only improves performance but also increases robustness over standard LoRA, underscoring the practical advantages of ShareLoRA in LLMs.

\paragraph{SVD Analysis of LoRA and ShareA Weights}

We conducted a Singular Value Decomposition (SVD) analysis on the weights of LLaMA 13B for both LoRA and ShareA, as shown in Figure~\ref{fig:SVD} in the Appendix. The results reveal distinct patterns in their singular value distributions across layers. LoRA weights exhibit a sharp decrease in singular values, indicating a concentration of information in a few dominant components. This could lead to specialization but might also increase the risk of overfitting. In contrast, ShareA weights show a smoother, more gradual decrease in singular values, suggesting a more balanced distribution of information among components. This balanced distribution contributes to ShareA's enhanced adaptability and generalization capability across different tasks. 

These findings provide insight into why ShareA may offer improved robustness and continue training performance compared to LoRA. The more uniform singular values distribution in ShareA suggests that it captures richer features, leading to better generalization across various domains.

\section{Conclusion}

In this paper, we introduce ShareLoRA, an optimization of the LoRA architecture that shares either the up or down projection across different layers. ShareLoRA significantly reduces the number of trainable parameters by at least half relative to the original LoRA and shows improved performance on fully converged datasets. Through extensive experimentation with NLU, NLG, and zero-shot tasks on models of varying scales, ShareLoRA demonstrates a strong balance between computational efficiency and robust performance. It consistently maintains high adaptability, strong robustness, and effective continual learning capabilities across diverse tasks and architectures.

\clearpage
\section{Limitation}
The limitations of ShareLoRA are  primarily in its convergence speed and practical applications. ShareAB and ShareB tend to converge more slowly compared to LoRA, though ShareA shows a convergence rate that is largely competitive with LoRA on smaller datasets, with only a slight lag on larger datasets. This indicates that ShareA is quite adept at easily converged datasets and effectively mitigating near-overfitting scenarios.\\
Regarding the practical application of GPUs, ShareLoRA introduces some complexities in the parallel training process on multiple GPUs. This is primarily due to the need for consistent synchronization of the Shared Module, once it is replicated across various GPUs at every computational step.

\bibliography{custom}
\bibliographystyle{acl_natbib}

\appendix
\clearpage

\begin{table*}[htbp]
\centering
\scalebox{0.85}{
\begin{tabularx}{1.15\textwidth}{l|c|c||l|c|c}
\toprule
\textbf{Method}           & \textbf{\# Params} & \textbf{MMLU} & \textbf{Method} & \textbf{\# Params} & \textbf{MMLU} \\ 
\midrule
LLaMA 7B (QLoRA)*              & 79.9M   & 38.8            & LLaMA 13B (QLoRA)*               & 125.2M   & 47.8  \\
LLaMA 7B (QLoRA)*              & 79.9M   & 39.96           & LLaMA 13B (QLoRA)*               & 125.2M   & 47.29 \\
LLaMA 7B (QLoRA)               & 79.9M   & $40.63\pm0.9$   & LLaMA 13B (QLoRA)                & 125.2M   & $47.13\pm0.9$ \\
LLaMA 7B (QShareA$_{qkv}$)     & 67.7M   & $40.63\pm0.5$   & LLaMA 13B (QShareA$_{qkv}$)      & 106.0M   & $\mathbf{47.36\pm0.7}$ \\ 
LLaMA 7B (QShareA)             & 44.6M   & $\mathbf{41.11\pm0.2}$   & LLaMA 13B (QShareA)              & 69.5M    & $47.17\pm0.8$\\ 
\bottomrule
\end{tabularx}
}
\caption{Performance comparison of LLaMA 7B and 13B with QLoRA and QShareA under the same configuration of \cite{dettmers2023qlora}, $*$ is similar experiment results collected from prior work \cite{xu2023parameterefficient}} \label{qShareLoRA}
\end{table*}

\begin{figure*}[ht!]
  \centering 
  \includegraphics[width=\textwidth]{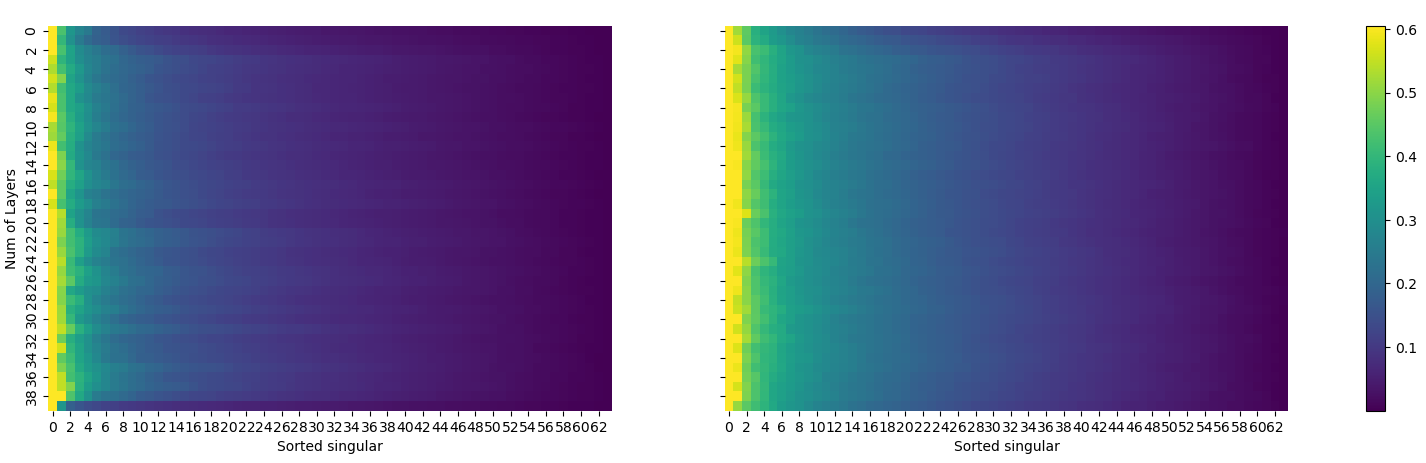} 
  \caption{Distribution of Singular Values for LLaMA 13B: SVD Decomposition Analysis of LoRA (left) and ShareA (right) across All Layers.}
  \label{fig:SVD}
\end{figure*}

\section{Hyperparameters}


In our study, we limits the extent of hyperparameter optimization in order to maintain consistency with prior research \cite{hu2021LoRA,dettmers2023qlora,mahabadi2021parameterefficient,eval-harness}, facilitating a direct comparison. Furthermore, we aims to investigate the behaviors of underfitting and overfitting across different scenarios using the LoRA and ShareLoRA approaches applied to various model size.

Specifically, under the current training setup, both LoRA and ShareLoRA exhibit signs of non-convergence when applied to the LLaMA 7B model. On the other hand, LoRA demonstrates clear overfitting when used with the LLaMA2 13B model, suggesting that the model training has gone beyond the point of optimal generalization.

For the models LLaMA 13B and LLaMA 2 7B, their performances are comparable. Both models reach a point of convergence and display fluctuations around this state, indicating that they are fully trained. It helps us understand the differing impacts of LoRA and ShareLoRA on these models under a set of reasonable training configurations.

The hyperparameter setting for RoBERTa is in Table~\ref{tab:training_details1} and for LLaMA are in Table~\ref{tab:parameters1} and \ref{tab:parameters2}.
The number of trainable parameters in Table~\ref{qShareLoRA}, should remain consistent between QLoRA and LoRA for LLaMA 7B and 13B in Table~\ref{MMLU}, as both models utilize BFloat16. However, the reduced number of trainable parameters is influenced by the implementation described in \cite{dettmers2023qlora}, which reduces the trainable parameters by half when quantizing to 4 bits. This is also reported the same by \cite{xu2023parameterefficient}, and we maintain this parameter count to ensure consistency.\\
We conducted five experiments with Roberta and GPT-2, and three experiments for all tasks related to LLaMA using different seeds. The results presented are all averages.

\section{LLaMA Performance Analysis}\label{llama_plot}



\begin{figure*}[ht]
  \centering 
  \includegraphics[width=0.95\linewidth]{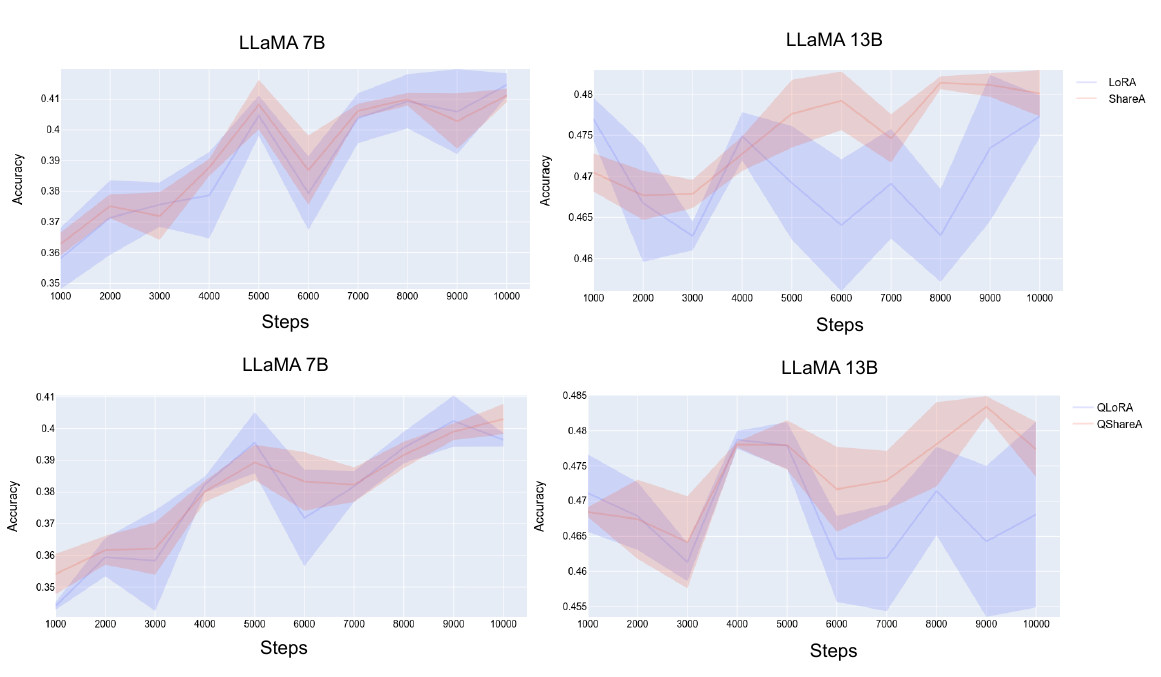} 
  \caption{LLaMA 7B \& 13B on LoRA / ShareA (upper) and on QLoRA / QShareA (down) MMLU Dev Performance with the standard deviation error distribution of different seeds}
  \label{fig:confidence}
\end{figure*}

\begin{figure*}[ht!]
  \centering 
  \includegraphics[width=0.9\textwidth]{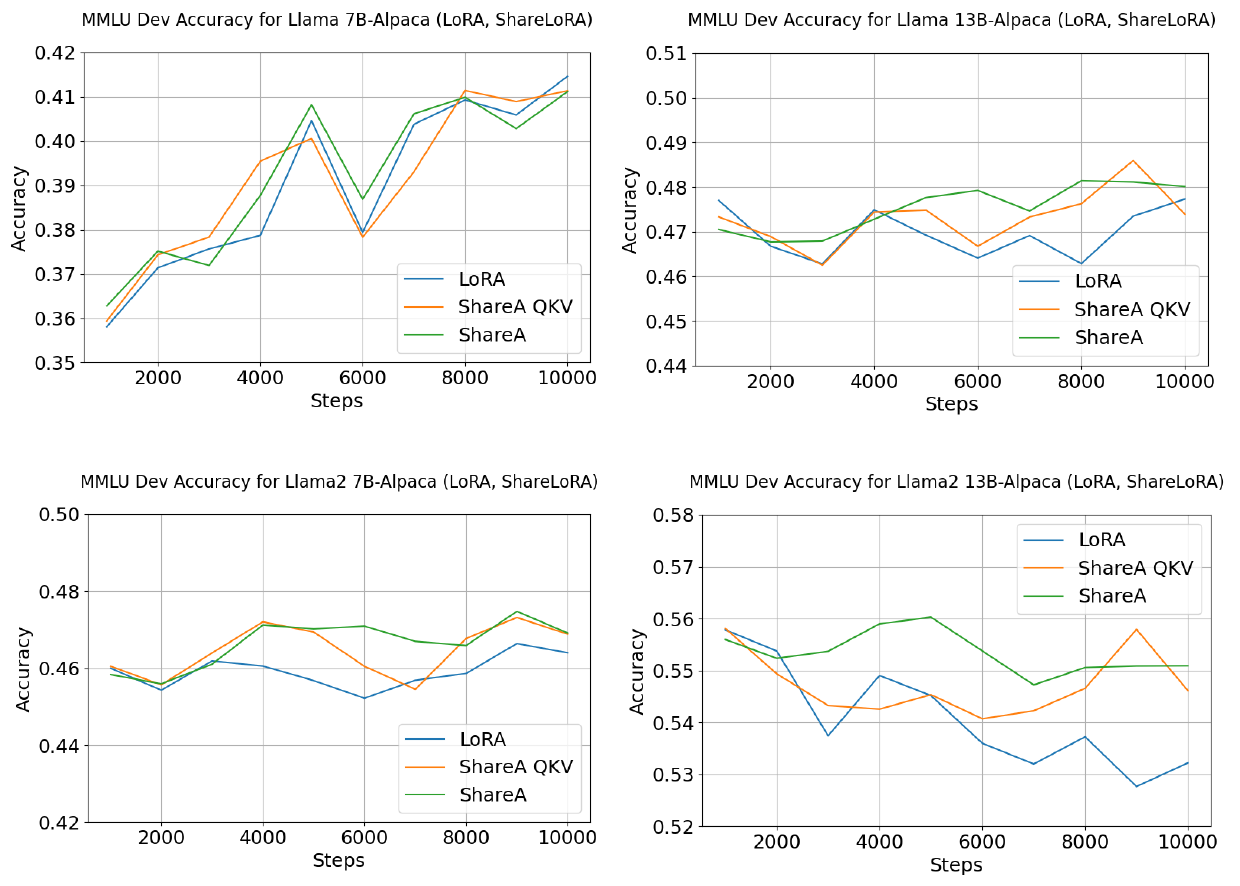} 
  \caption{Average Performance Plot for Various LLaMA Models on the Alpaca-MMLU Dev Dataset}
  \label{fig:plot}
\end{figure*}

In Figures \ref{fig:confidence} and \ref{fig:plot}, we present the Dev Set performance changes for both LLaMA and LLaMA2 models, ranging from 7B to 13B, to observe the differences in performance over steps. The results demonstrate that ShareA and ShareA$_{qkv}$ configurations offer several advantages over their counterparts, as discussed in Section \ref{QKV}.

For both the 7B and 13B models, ShareA and ShareA$_{qkv}$ configurations maintain higher average accuracy compared to the traditional LoRA setup. Specifically, ShareA demonstrates consistent performance improvements, particularly in the stability of accuracy over different steps. This indicates that ShareA is more robust and less prone to fluctuations compared to LoRA.

The robustness of ShareLoRA extends to quantized models. Table~\ref{qShareLoRA} shows that QShareA (QLoRA with ShareA) maintains strong performance even with substantial parameter reduction. In the case of LLaMA 7B, QShareA achieves an MMLU score of 41.11, surpassing QLoRA's score of 40.63. This trend continues with larger models: for LLaMA 13B, QShareA slightly outperforms QLoRA with scores of 47.17 and 47.13 respectively, while using significantly fewer parameters. These performance gains are consistently observed across different model sizes, including LLaMA2 7B and LLaMA 13B, highlighting ShareLoRA's broad applicability and scalability.

The analysis in Figure \ref{fig:confidence} further enriches our results by incorporating confidence intervals which map the performance stability of LoRA, QLoRA, ShareA, and QShareA. From these plots, it is evident that while LoRA occasionally outperforms QLoRA, the overall performance trends of LoRA and QLoRA are closely aligned in LLaMA 7B. In particular, for the LLaMA 13B, the performance of ShareA and QShareA after 5000 steps is completely superior than LoRA and QLoRA. It is crucial to highlight that both LoRA and QLoRA display larger fluctuations in performance compared to ShareA and QShareA, underscoring a potentially greater variability in model outcomes across different experimental seeds.

\begin{figure*}[ht!]
  \centering 
  \includegraphics[width=\textwidth]{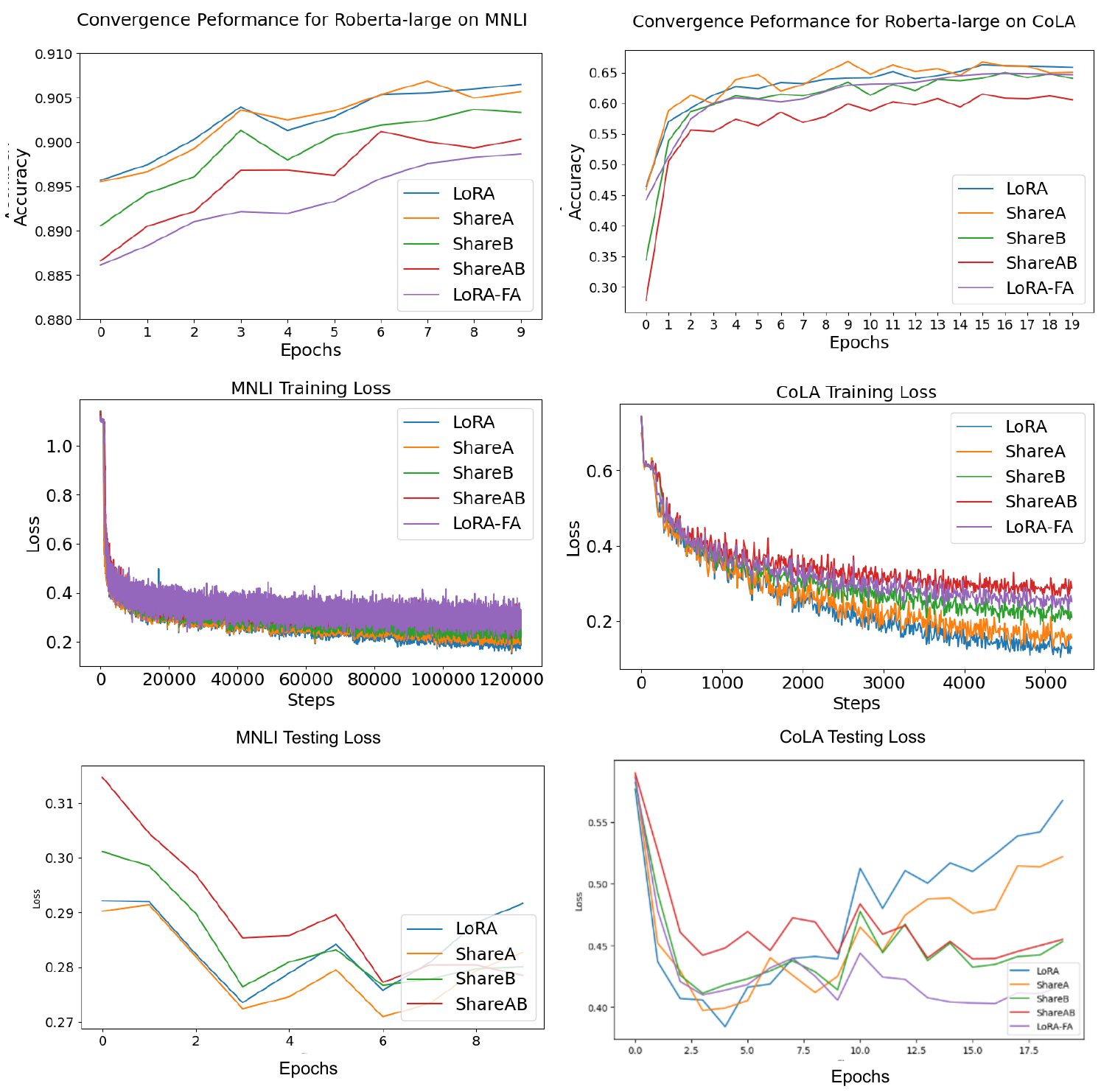} 
  \caption{Convergence Performance for MNLI and CoLA datasets}
  \label{fig:convergence}
\end{figure*}

\section{Convergence Analysis} 
In Figure \ref{fig:convergence}, we analyze the convergence trends across both the MNLI and CoLA datasets for the RoBERTa-large model, demonstrating differing behaviors among the sharing strategies and others. Notably, while ShareA begins with slightly lower performance compared to LoRA, it progressively matches LoRA's accuracy on the MNLI dataset. ShareB and ShareAB, in contrast, consistently underperform relative to both LoRA and ShareA. This pattern is similarly observed with the CoLA dataset, where ShareA's performance is robust, closely competing with LoRA. Both ShareB and ShareAB are worse than LoRA alone.

At the same time, LoRA-FA only reaches performance levels comparable to ShareB, lagging behind both ShareA and LoRA. This suggests that ShareA not only sustains competitive convergence capabilities but also outperforms LoRA-FA in terms of robustness and eventual alignment with LoRA's top performance.

In term of training loss, all models exhibit a similar declining trend over the training epochs. However, ShareA distinguishes itself by slightly lagging behind LoRA initially in terms of speed of convergence but substantial surpassing both ShareB and LoRA-FA overall. This differential suggests that ShareA offers a balanced approach, effectively managing a slower initial convergence for consistent long-term gains.




\begin{table*}[ht]
\centering
\scalebox{0.9}{
\begin{tabular}{@{}clcccccccc@{}}
\toprule
\textbf{Method}    & \textbf{Dataset} & \textbf{MNLI} & \textbf{SST-2} & \textbf{MRPC} & \textbf{CoLA} & \textbf{QNLI} & \textbf{QQP} & \textbf{RTE} & \textbf{STS-B} \\ \midrule
&Optimizer                            & \multicolumn{8}{c}{AdamW} \\
&Warmup Ratio                         & \multicolumn{8}{c}{0.06} \\
&LR Schedule                          & \multicolumn{8}{c}{Linear}\\ \midrule 
&Batch Size (per device)              & 16           & 16            & 16            & 32            & 32            & 16            & 32           & 16            \\
&\# Epochs                            & 30           & 60            & 30            & 80            & 25            & 25            & 80           & 40            \\
RoBERTa base & Learning Rate                        & 5E-04        & 5E-04         & 4E-04         & 4E-04         & 4E-04         & 5E-04         & 5E-04        & 4E-04         \\
ShareLoRA&LoRA Config.          & \multicolumn{8}{c}{$r_q = r_v = 8$} \\
&LoRA $\alpha$                       & \multicolumn{8}{c}{8} \\
&Max Seq. Len.                    & \multicolumn{8}{c}{512} \\ 
&seed                    & \multicolumn{8}{c}{0,1,2,3,4} \\ \midrule

&Batch Size  (per device)             & \multicolumn{8}{c}{4} \\         
&\# Epochs                            & 10           & 10            & 20            & 20            & 10            & 20            & 20           & 10            \\
RoBERTa large   &Learning Rate                        & 3E-04        & 4E-04         & 3E-04         & 2E-04         & 2E-04         & 3E-04         & 4E-04        & 2E-04         \\
ShareLoRA $\dag$  &LoRA Config.          & \multicolumn{8}{c}{$r_q = r_v = 8$} \\
&LoRA $\alpha$                       & \multicolumn{8}{c}{8} \\
&Max Seq. Len.                    & \multicolumn{8}{c}{512} \\ 
&seed                    & \multicolumn{8}{c}{0,1,2,3,4} \\
\bottomrule
\end{tabular}
}
\caption{Configuration and training details for RoBERTa base LoRA on different datasets.}
\label{tab:training_details1}
\end{table*}

\begin{table*}[ht]
\centering
\scalebox{0.9}{
\begin{tabular}{@{}clcccccccc@{}}
\toprule
\textbf{Dataset} & \textbf{E2E Challege} \\ \midrule
Optimizer                            & AdamW \\
Weight Decay                         & 0.01 \\
Dropout Prob                         & 0.1\\ 
Batch Size (per device)              & 8 \\
\# Epochs                            & 5 \\
Warmup Steps                         & 500   \\
Learning Rate Schedule               & Linear \\
Label Smooth                         & 0.1 \\
Learning Rate                        & 0.002 \\ 
Adaptation                           & $r_q = r_v = 4$  \\ \midrule
LoRA $\alpha$                        & 32 \\
Beam Size.                           & 10 \\ 
Length Penalty                       & 0.9  \\
no repeat ngram size                 & 4  \\
\bottomrule
\end{tabular}
}

\caption{Configuration and training details for GPT-2 LoRA on E2E Challenge}
\label{tab:training_details2}
\end{table*}

\begin{table*}[ht]
\centering
\scalebox{0.9}{
\begin{tabular}{@{}ccccccccc@{}}
\toprule
\textbf{Model} &\textbf{Parameters}  & \textbf{Batch size} & \textbf{LR} & \textbf{Steps} & \textbf{Source Length} & \textbf{Target Length} & \textbf{LoRA r}  & \textbf{LoRA $\alpha$} \\ 
\midrule
LLaMA1 \& 2  &7B                             & 16                  & 2e-4        & 10000          & 384                    & 128     & 64    & 16                  \\
LLaMA1 \& 2 &13B                            & 16                  & 2e-4        & 10000           & 384                    & 128    & 64    & 16                  \\
LLaMA3 \& 3.1 &8B                            & 16                  & 2e-5        & 5000           & 384                    & 128    & 64    & 16                  \\
\bottomrule
\end{tabular}}
\caption{Training hyperparameters for LLaMA and QLLaMA.}
\label{tab:parameters1}
\end{table*}

\begin{table*}[ht]
\centering
\begin{tabular}{@{}ccccc@{}}
\toprule
\textbf{Parameters}            & \textbf{MMLU Source Length} & \textbf{Temperature} & \textbf{Top P} & \textbf{Beam size} \\ \midrule
7B                             & 2048                  & 0.7         & 0.9           & 1                                     \\
13B                            & 2048                  & 0.7        & 0.9           & 1                                     \\
\bottomrule
\end{tabular}
\caption{Evaluation hyperparameters for LLaMA and QLLaMA.}
\label{tab:parameters2}
\end{table*}

\begin{figure*}[ht!]
  \centering 
  \includegraphics[width=0.7\textwidth]{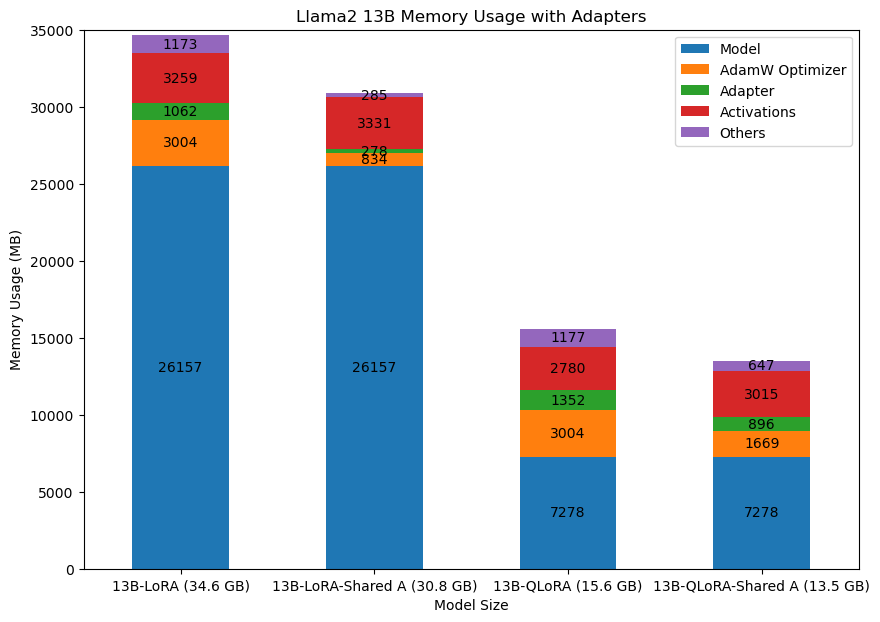} 
  \caption{Memory Consumption required for LLaMA2 13B.}
  \label{fig:memory_size}
\end{figure*}

\section{Memory Footprint}
We utilizes float32 for QLoRA modules to enhance the performance of quantized models, while bfloat16 is employed for LoRA fine-tuning. We employ the standard AdamW optimizer with a batch size of 1, a sequence length of 512, and do not use gradient checkpointing. 

The chart in Figure ~\ref{fig:memory_size} depicts memory usage across four configurations of the Llama2 13B model: LoRA, LoRA-Shared A, QLoRA, and QLoRA-Shared A, highlighting the impact of model scaling and adaptations on resource needs. It shows a memory reduction of 3.8 GB when using LoRA-Shared A compared to the LoRA configuration, and a further savings of 2.1 GB with QLoRA-Shared A compared to QLoRA. LoRA-Shared operates independently from QLoRA strategies, thereby reducing memory usage further without interfering with LoRA or QLoRA configurations.

\end{document}